\documentclass[10pt,twocolumn,letterpaper]{article}

\usepackage{cvpr}
\usepackage{times}
\usepackage{epsfig}
\usepackage{graphicx}
\usepackage{amsmath}
\usepackage{amssymb}

\graphicspath{ {figs/} }

\usepackage{booktabs}
\usepackage{makecell}
\usepackage{tabu,multirow}
\usepackage{pbox}
\usepackage{multicol}
\usepackage{hhline}
\usepackage{subcaption}
\usepackage{enumitem}
\usepackage[normalem]{ulem}

\usepackage[dvipsnames]{xcolor}

\newcommand{\ours}[0]{Graph-Based Global Reasoning Networks\xspace}

\newcommand{\oursunitlong}[0]{Global Reasoning unit\xspace}
\newcommand{\oursunit}[0]{\oursunitshort unit\xspace}
\newcommand{\oursunits}[0]{\oursunitshort units\xspace}
\newcommand{\oursunitshort}[0]{GloRe\xspace}
\usepackage{xspace}
\newcommand{\head}[1]{{\smallskip\noindent\bf{#1}}}

\newcommand{\vZ}[0]{{\mathbf{Z}}}

\newcommand{\G}[0]{{V}}
\newcommand{\vx}[0]{{\mathbf{x}}}
\newcommand{\vy}[0]{{\mathbf{y}}}
\newcommand{\vz}[0]{{\mathbf{z}}}

\newcommand{\vb}[0]{{\mathbf{b}}}

\newcommand{\vd}[0]{{\mathbf{d}}}

\newcommand{\vg}[0]{{\mathbf{v}}}

\makeatletter
\DeclareRobustCommand\onedot{\futurelet\@let@token\@onedot}
\def\@onedot{\ifx\@let@token.\else.\null\fi\xspace}
\def\eg{e.g\onedot} 
\def\ie{\emph{i.e}\onedot}

\def\etal{\emph{et~al}\onedot}

\makeatother
\usepackage[pagebackref=true,breaklinks=true,letterpaper=true,colorlinks,bookmarks=false]{hyperref}

\cvprfinalcopy %

\def\cvprPaperID{750} %

\ifcvprfinal\fi
\begin{document}

\title{Graph-Based Global Reasoning Networks}

\author{
  Yunpeng Chen$^\dagger$$^\ddagger$, Marcus Rohrbach$^\dagger$, Zhicheng Yan$^\dagger$, Shuicheng Yan$^\ddagger$$^\flat$, Jiashi Feng$^\ddagger$, Yannis Kalantidis$^\dagger$ \\
  $^\dagger$Facebook Research, $^\ddagger$National University of Singapore, $^\flat$Qihoo 360 AI Institute \\
}

\maketitle
\begin{abstract}
\color{black}

Globally modeling and reasoning over relations between regions can be beneficial for many computer vision tasks on both images and videos. Convolutional Neural Networks (CNNs) excel at modeling local relations by convolution operations, but they are typically inefficient 
at capturing global relations between distant regions and require stacking multiple convolution layers.
In this work, we propose a new approach for \emph{reasoning globally} in which a set of features are globally aggregated over the coordinate space and then projected to an interaction space where relational reasoning can be efficiently computed. After reasoning, relation-aware features are distributed back to the original coordinate space for down-stream tasks. We further present a highly efficient instantiation of the proposed approach and introduce the \oursunitlong (\oursunit) that implements the coordinate-interaction space mapping by weighted global pooling and weighted broadcasting, and the relation reasoning via graph convolution on a small graph in interaction space.
The proposed \oursunit is lightweight, end-to-end trainable and can be easily plugged into existing 
CNNs for a wide range of tasks. Extensive experiments show our \oursunit can consistently boost the performance of state-of-the-art backbone architectures,  including ResNet~\cite{he2016deep,he2016identity}, ResNeXt~\cite{xie2017aggregated}, SE-Net~\cite{hu2017} and DPN~\cite{chen2017dual}, for both 2D and 3D CNNs, on image classification, semantic segmentation and video action recognition task. 

\color{Purple}
\end{abstract}

\color{black}

\section{Introduction}
\label{sec:introduction}

\color{black}

\begin{figure}[t]
\centering
\resizebox{\columnwidth}{!}{
	\includegraphics[]{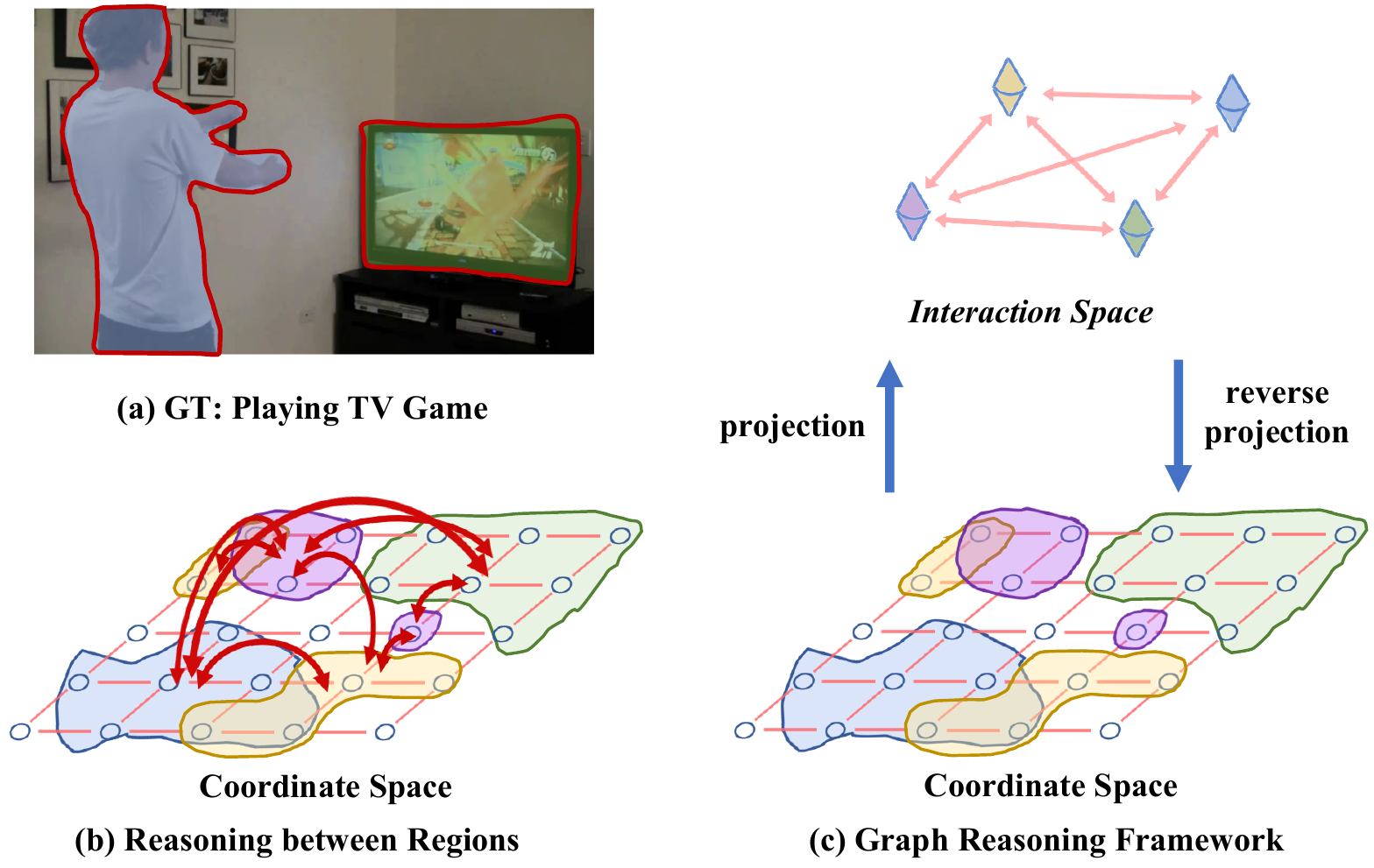}
}
\caption{
Illustration of our main idea. Aiming at capturing relations between arbitrary regions over the full input space (shown in different colors), we propose a novel approach for reasoning globally (shown in Fig. (c)). Features from the colored regions in coordinate space are projected into nodes in \emph{interaction space}, forming a fully-connected graph. After reasoning over the graph, node features are projected back to the coordinate space. 
}

\label{fig:teaser}
\end{figure}

Relational reasoning between distant regions of arbitrary shape is crucial for many computer vision tasks like image classification~\cite{chen2004image}, segmentation~\cite{yang2018denseaspp, zhao2017pyramid} and action recognition~\cite{wang2017non}. %
Humans can easily understand the relations among different regions of an image/video, as shown in Figure~\ref{fig:teaser}(a). However, deep CNNs cannot capture such relations without stacking multiple convolution layers, since an individual layer can only capture information locally. This is very inefficient, since relations between distant regions of arbitrary shape on the feature map can only be captured by a near-top layer with a sufficiently large receptive field to cover all the regions of interest. 
For instance, in ResNet-50~\cite{he2016deep} with 16 residual units, the receptive field is gradually increased to cover the entire the image of size $224 \times 224$ at 11th unit (the near-end of Res4).
To solve this problem, we propose a unit to directly perform global relation reasoning by projecting features from regions of interest to an interaction space and then distribute back to the original coordinate space.
In this way, relation reasoning can be performed in early stages of a CNN model.
Specifically, rather than relying solely on convolutions in the coordinate space to implicitly model and communicate information among different regions,
we propose to construct a \emph{latent interaction space} where global reasoning can be performed directly, as shown in Figure~\ref{fig:teaser}(c). 
Within this interaction space, a set of regions that share similar semantics are represented by a single feature, instead of a set of scattered coordinate-specific features from the input. Reasoning the relations of multiple different regions is thus simplified to modeling those between the corresponding features in the interaction space, as shown on the top of Figure~\ref{fig:teaser}(c).
We thus build a graph connecting these features within the interaction space and perform relation reasoning over the graph. After the reasoning, the updated information is then projected back to the original coordinate space for down-streaming tasks. 
Accordingly, we devise a \emph{\oursunitlong} (\emph{\oursunitshort}) to efficiently implement the coordinate-interaction space mapping process by weighted global pooling and weighted broadcasting, as well as the relation reasoning by graph convolution~\cite{kipf2016semi}, which is differentiable and also end-to-end trainable.

\vspace{1mm}
Different from the recently proposed Non-local Neural Networks (NL-Nets)~\cite{wang2017non} and Double Attention Networks~\cite{chen2018a2nets} which only focus on delivering information and rely on convolution layers for reasoning, our proposed model is able to directly 
reason on relations over regions. 
Similarly, Squeeze-and-Extension Networks (SE-Nets)~\cite{hu2017} only focus on incorporating image-level features via global average pooling, leading to an interaction graph containing only one node. It is not designed for regional reasoning as our proposed method.
Extensive experiments show that inserting our \oursunitshort can consistently boost performance of state-of-the-art CNN architectures on diverse tasks including image classification, semantic segmentation and video action recognition.

\vspace{2mm}
\noindent Our contributions are summarized below:
\vspace{-1mm}
\begin{itemize}[leftmargin=*]
 \item We propose a new approach for \emph{reasoning globally} by projecting a set of features that are globally aggregated over the coordinate space into an interaction space where relational reasoning can be efficiently computed. After reasoning, relation-aware features are distributed back to the coordinate space for down-stream tasks.
\item We present the \oursunitlong (\oursunit) a highly efficient instantiation of the proposed approach that implements the coordinate-interaction space mapping by weighted global pooling and weighted broadcasting, and the relation reasoning via graph convolution in the interaction space.
\item We conduct extensive experiments on a number of datasets and show the \oursunitlong can bring consistent performance boost for a wide range of backbones including ResNet, ResNeXt, SE-Net and DPN, for both 2D and 3D CNNs, on image classification, semantic segmentation and video action recognition task. 

\end{itemize}

\noindent

\color{black}

\section{Related Work}
\label{sec:related}
\head{Deep Architecture Design}.
Research on deep architecture design focuses on building more efficient convolution layer topologies, aiming at alleviating optimization difficulties or increasing  efficiency of backbone architectures.
Residual Networks (ResNet)~\cite{he2016deep,he2016identity} and DenseNet~\cite{huang2017densely} are proposed to alleviate the optimization difficulties of deep neural networks. DPN~\cite{chen2017dual} combines benefits of these two networks with further improved performance. Xception~\cite{chollet2017xception}, MobileNet~\cite{howard2017mobilenets,sandler2018inverted}, and ResNeXt~\cite{xie2017aggregated} use grouped or depth-wise convolutions to reduce the computational cost. Meanwhile, reinforcement learning based methods~\cite{zoph2016neural} try to automatically find the network topology in a predefined search space. All these methods, though effective, are built by stacking convolution layers and thus suffer low-efficiency of convolution operations on reasoning between disjoint or distant regions. In this work we propose an auxiliary unit that can overcome this shortage and bring significant performance gain for these networks. 
\head{Global Context Modeling}.
Many efforts try to overcome the limitation of local convolution operators by introducing global contexts.  PSP-Net~\cite{zhao2017pyramid} and DenseASPP~\cite{yang2018denseaspp} combine multi-scale features to effectively enlarge the receptive field of the convolution layers for segmentation tasks. Deformable CNNs~\cite{dai2017deformable} achieve the similar outcome by further learning offsets for the convolution sampling locations. Squeeze-and-extension Networks~\cite{hu2017} (SE-Net) use global average pooling to incorporate an image-level descriptor at every stage. Nonlocal Networks~\cite{wang2017non}, self-attention Mechanism~\cite{vaswani2017attention} and Double Attention Networks (A$^2$-Net)~\cite{chen2018a2nets} try to deliver long-range information from one location to another. Meanwhile, bilinear pooling~\cite{lin2015bilinear} extracts image level second-order statistics to complement the convolution features. Although we also incorporate global information, in the proposed approach we go one step further and perform higher-level reasoning on a graph of the relations between disjoint or distant regions as shown in Figure~\ref{fig:teaser}(b).

\head{Graph-based Reasoning}. 
Graph-based methods have been very popular in recent years and shown to be an efficient way of relation reasoning. CRFs~\cite{chandra2017dense} and random walk networks~\cite{bertasius2017convolutional} are proposed based on the graph model for effective image segmentation. Recently,  Graph Convolution Networks (GCN)~\cite{kipf2016semi} are proposed for semi-supervised classification, and Wang \etal~\cite{wang2018videos} propose to use GCN to capture relations between objects in video recognition tasks, where objects are detected by an object detector pre-trained on extra training data. In contrast to~\cite{wang2018videos}, we adopt the reasoning power of graph convolutions to build a generic, end-to-end trainable module for reasoning between disjoint and distant regions, regardless of their shape and without the need for object detectors or extra annotations.

\section{Graph-based Global Reasoning}
\label{sec:method}
\color{black}

In this section, we first provide an overview of the proposed \oursunitlong, the core unit to our graph-based global reasoning network, and introduce the motivation and rationale for its design. We then describe its architecture in details. Finally, we elaborate on how to apply it for several different computer vision tasks. 

Throughout this section, for simplicity, all figures are plotted  based on 2D (image) input tensors. A graph $G =(\mathcal{V}, \mathcal{E}, A)$ is typically defined by its nodes  $\mathcal{V}$, edges $\mathcal{E}$ and adjacent matrix $A$ describing the edge weights. In the following, we interchangeably use $A$ or $G$ to refer to a graph defined by $A$. 
\subsection{Overview}
\label{sec:method:overview}
Our proposed \oursunit is motivated by overcoming the intrinsic limitation of convolution operations for modeling global relations. For an input feature tensor $X \in \mathbb{R}^{L \times C}$, with $C$ being the feature dimension and $L = W \times H$ locations, standard convolutional layers process inputs w.r.t.\ the regular grid coordinates $\Omega = \{1, \ldots,H\} \times \{1,\ldots, W\}$ to extract features. Concretely, the convolution is performed over a regular nearest neighbor graph defined by an adjacent matrix $A \in \mathbb{R}^{L \times L}$ where $A_{ij}=1$ if regions $i$ and $j$ are spatially adjacent, and otherwise $A_{ij}=0$.
The edges of the graph  encode spatial proximity and its node stores the feature for that location as shown on the bottom of Figure~\ref{fig:teaser}(c). Then the output features of such a convolution layer are computed as $Y=AXW$ where $W$ denotes parameters of the convolution kernels.  A single convolution layer can capture local relations covered by the convolution kernel (\ie, locations connected over the graph $A$). But capturing relations among  disjoint and distant regions of arbitrary shape requires stacking multiple such convolution layers, which is highly inefficient. Such a drawback increases the difficulty and cost of global reasoning for CNNs.

To solve this problem, we propose to first project the features $X$ from the coordinate space $\Omega$ to the features $V$ in a latent interaction space $\mathcal{H}$, where  each set of disjoint regions can be represented by a single feature instead of a bunch of features at different locations. Within the interaction space $\mathcal{H}$, we can build a new \emph{fully-connected} graph $A_g$, where each node stores the new feature as its state.  In this way, the relation reasoning is simplified as modeling the interaction between pairs of nodes over a smaller graph $A_g$ as shown on the top of the Figure~\ref{fig:teaser}(c). 

\begin{figure}[t]
\centering
\resizebox{1.0\columnwidth}{!}{
	\includegraphics[]{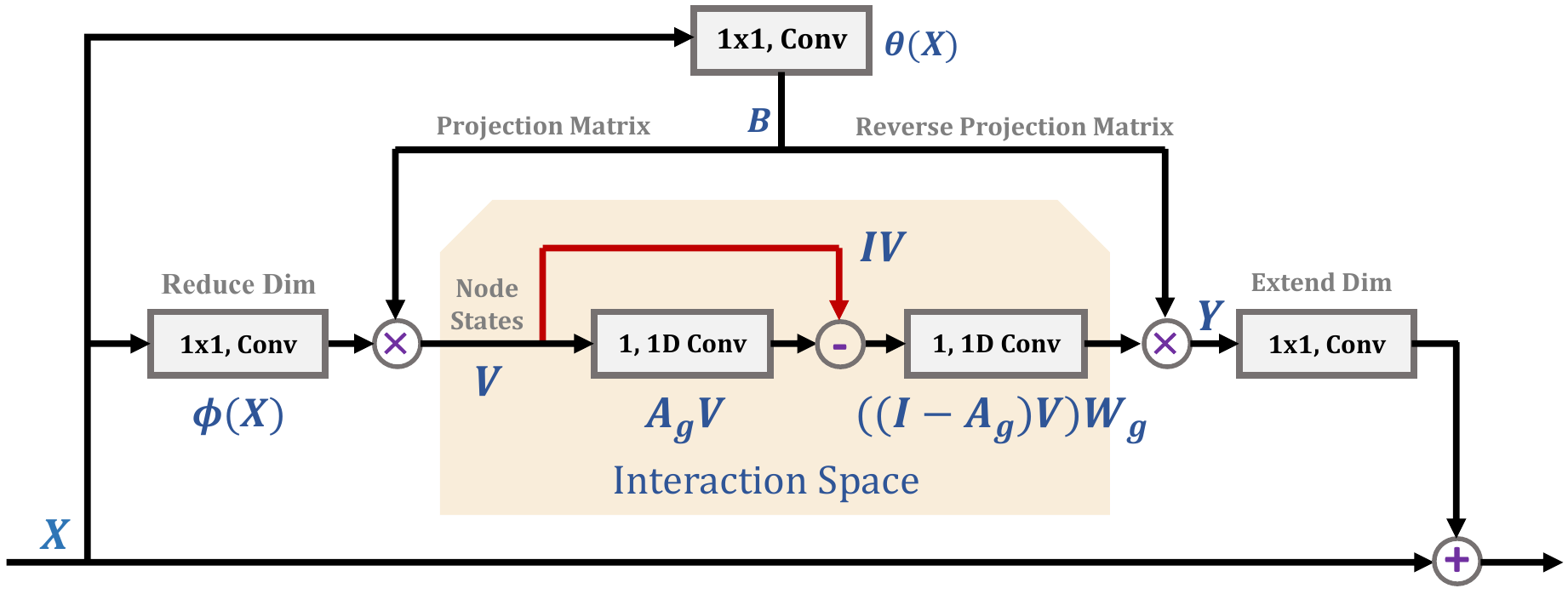}
}
\caption{Architecture of the proposed \oursunitlong. It consists of five convolutions, two for dimension reduction and expansion (the left and right most ones) over input features $X$ and output $Y$, one for generating the bi-projection $B$ between the coordinate and latent interaction spaces (the top one), and two for global reasoning based on the graph $A_g$ in the interaction space (the middle ones). Here $V$ encodes the regional features as graph nodes and $W_g$ denotes parameters for the graph convolution.
}
\label{fig:method:our-lock}
\end{figure}

Once we obtain the feature for each node of graph $A_g$, we apply a general graph   convolution  to model and reason about  the contextual relations between each pair of nodes. After that, we perform a reverse projection to transform the resulting features (augmented with relation information) back to the original coordinate space, providing complementary features for the following layers to learn better task-specific  representations. Such a three-step process is conceptually depicted in Figure~\ref{fig:teaser}(c). To implement this process, we propose a highly efficient unit, termed \oursunit, with its architecture outlined in Figure~\ref{fig:method:our-lock}. 

In the following subsections, we describe each step of the proposed \oursunit in detail.

\subsection{From Coordinate Space to Interaction Space}
\label{sec:method_projections}

The first step is to find the projection function $f(\cdot)$ that maps original features to the interaction space $ \mathcal{H}$. Given a set of input features  $X\in\mathbb{R}^{L \times C}$, we aim to learn the projection function such that the new features $V=f(X) \in \mathbb{R}^{N \times C}$ in the interaction space are more friendly for global reasoning over disjoint and distant regions. 
Here $N$ is the number of the features (nodes) in the interaction space. 
Since we expect to directly reason over a set of regions, as shown in Figure~\ref{fig:teaser}(b), 
we formulate the projection function as a linear combination (\emph{a.k.a} weighted global pooling) of original features such that the new features can aggregate information from multiple regions. In particular, each new feature is generated by  
\begin{equation}
\vg_i = {\vb_i}{X}  = \sum_{\forall j}b_{ij}\vx_j, 
\label{eqn:global_projection}
\end{equation}
with learnable projection weights $B=[\vb_1, \cdots, \vb_{N}] \in \mathbb{R}^{N \times L}$, $\vx_j \in \mathbb{R}^{1 \times C}$, $\vg_i \in \mathbb{R}^{1 \times C}$. 

We note that the above equation gives a more generic formulation than an existing method~\cite{wang2018videos}, where an object detector pre-trained on an extra dataset is adopted to determine $\vb_i$, \ie $b_{ij} = 1$ if $j$ is inside the object box, and $b_{ij} = 0$ if it is outside the box. Instead of using extra annotation and introducing a time-consuming object detector to form a binary combination, we propose to use convolution layers to directly generate $\vb_i$ (we use one convolution layer in this work).

In practice, to reduce input dimension and enhance capacity of the projection function, we implement the function $f(X)$ as $f(\phi(\mathbf{X} ; W_{\phi}))$ and $B = \theta(\mathbf{X} ; W_{\theta})$. We model $\phi(\cdot)$ and $\theta(\cdot)$ by two convolution layers as shown in Figure~\ref{fig:method:our-lock}. $W_\phi$ and $W_\theta$ are the learnable convolutional kernel of each layer. The benefits of directly using the output of a convolution layer to form the $\vb_i$ include the following aspects. 1) The convolution layer is end-to-end trainable. 2) Its training does not require any object bounding box as \cite{wang2018videos}. 3) It is simple to implement and faster in speed. 4) It is more generic since the convolution output can be both positive and negative, which linearly fuses the information in the coordination space. %
\subsection{Reasoning with Graph Convolution}
\label{sec:method_graph}
After projecting the features from coordinate space into the interaction space, we have graph where each node contains feature descriptor. Capturing relations between arbitrary regions in the input is now simplified to capturing interactions between the features of the corresponding nodes.

There are several possible ways of capturing the relations between features in the new space. The most straightforward one would be to concatenate the features as input and use a small neural network to capture inter-dependencies, like the one proposed in~\cite{santoro2017simple}. However, even a simple relation network is computationally expensive and concatenation destroys the pair-wise correspondence along the feature dimension. 
Instead, we propose treating the features as nodes of a fully connected graph, propose to reason on the fully connected graph by learning edge weights that correspond to interactions of the underlying globally-pooled features of each node. To that end, we adopt the recently proposed graph convolution~\cite{kipf2016semi}, a highly efficient, effective and differentiable module.

In particular, let $G$ and $A_g$ %
denote the $N \times N$ node adjacency matrix for diffusing information across nodes, and let $W_g$ denote the state update function. A single-layer graph convolution network is defined by Eqn.~\eqref{eqn:gcn:main}, where the adjacency matrix $A_g$ is randomly initialized and \textit{learned} by gradient decent during training, together with the weights. 
The identity matrix serves as a shortcut connection that alleviates the optimization difficulties. The graph convolution~\cite{kipf2016semi,li2018deeper} is formulated as
\begin{equation}
\vZ = G \G W_g = ((I-A_g)\G) W_g.
\label{eqn:gcn:main}
\end{equation}

\begin{figure}[t]
\centering
\begin{subfigure}{\columnwidth}
  \centering
  \includegraphics[width=1.0\linewidth]{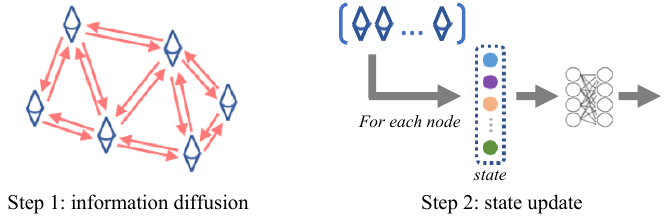}
  \caption{Graph Convolution seen as two steps.}
  \label{fig:method:gcn-two-step:a}
\end{subfigure}%
\\
\begin{subfigure}{\columnwidth}
  \centering
  \includegraphics[width=1.0\linewidth]{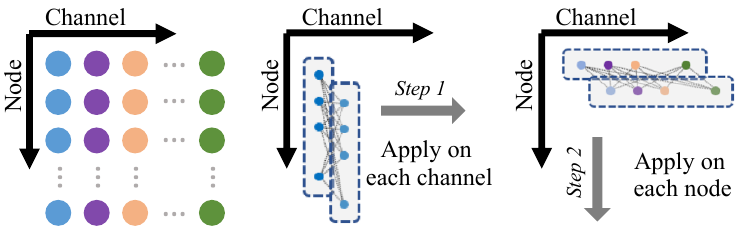}
  \caption{$\mathrm{GCN}(X) = \mathrm{Conv1D}( \mathrm{Conv1D}(X)^T)^T$}
  \label{fig:method:gcn-two-step:b}
\end{subfigure}
\caption{Relation reasoning through graph convolution. (a) An intuitive explanation of graph convolution. (b) Implementation of Graph Convolution using two-direction 1D convolutions.}%
\label{fig:method:gcn-two-step}
\end{figure}

The first step of the graph convolution performs Laplacian smoothing~\cite{li2018deeper}, propagating the node features over the graph. 
During training, the adjacent matrix learns edge weights that reflect the relations between the underlying globally-pooled features of each node. If, for example, two nodes contain features that focus on the eyes and the nose, learning a strong connection between the two would strengthen the features for a possible downstream ``face'' classifier. 
After information diffusion, each node has received all necessary information and its state is updated through a linear transformation. 
This two step process is conceptually visualized in Figure~\ref{fig:method:gcn-two-step}(a). In Figure~\ref{fig:method:gcn-two-step}(b), we show the implementation of this two step process and the graph convolution via two 1$D$ convolution layers along different directions, \ie channel-wise and node-wise. 

\subsection{From Interaction Space  to Coordinate Space}

To make the above building block compatible with existing CNN architectures, the last step is to project the output features back to the original space after the relation reasoning. In this way, the updated features from reasoning can be utilized by the following convolution layers to make better decisions. This reverse projection is very similar to the projection in the first step.

Given the node-feature matrix $Z\in \mathbb{R}^{N \times C}$, we aim to learn a mapping function that can transform  the features to $Y\in R^{L \times C}$ as follows:
\begin{equation}
Y = g(Z).
\label{eqn:bilinear}
\end{equation}
Similar to the first step, we adopt linear projection to formulate $g(Z)$:
\begin{equation}
\vy_i = {\vd_i}{Z}  = \sum_{\forall j}d_{ij}\vz_j.
\label{eqn:global_reverse-projection}
\end{equation}
The above projection is actually performing feature diffusion. The feature 
$\vz_j$ of node $j$ is assigned to $\vy_i$ weighted by a scalar $d_{ij}$.
These weighs form the dense connections from the semantic graph to the grid map. Again, one can force the weighted connections to be binary masks
or can simply use a shallow network to generate these connections. In our work, we use a single convolution layer to predict these weights. In practice, we find that we can reuse the projection generated in the first step to reduce the computational cost without producing any negative effect upon the final accuracy. In other words, we set $D = B^\top$.

The right most side of Figure~\ref{fig:method:our-lock} shows the detailed implementation. In particular, the information from the graph convolution layer is projected back to the original space through the weighted broadcasting in Eqn.~\eqref{eqn:global_reverse-projection}, where we reuse the output from the top convolution layer as the weight. Another convolution layer is attached after migrating the information back to the original space for dimension expansion, so that the output dimension can match the input dimension forming a residual path.

\subsection{Deploying the Global Reasoning Unit}
\label{sec:grn_unit}

The core processing of the proposed \oursunitlong happens after flattening all dimensions referring to locations. It therefore straightforwardly applies to 3D (\eg spatio-temporal) or 1D (\eg temporal or any one-dimensional) features by adapting the dimensions of the three convolutions that operate in the coordinate space and then flattening the corresponding dimensions. For example, in the 3D input case, the input is a set of frames and $L = H \times W \times T$, where $H, W$ are the spatial dimensions and $T$ is the temporal dimension, \ie the number of frames in the clip. In this case, the three $1 \time 1$ convolutional layers shown in Figure~\ref{fig:method:our-lock} will be replaced by $1 \times 1 \times 1$ convolutions.

In practice, due to its residual nature, the proposed \oursunitlong can be easily incorporated into a large variety of existing backbone CNN architectures. It is light-weight and can therefore be inserted one or multiple times throughout the network, reasoning global information at different stages and complementary to both shallow and deeper networks. Although the latter can in theory capture such relations via multiple stacked convolutions, we show that adding one or more of the proposed \oursunitlong increases performance for downstream tasks even for very deep networks. In the following section, we present results from different instantiations of \ours with one or multiple \oursunitlong at different stages, describing the details and trade-offs in each case. We will refer to networks with at least one \oursunitlong as \emph{\ours}.

\color{black}

\section{Experiments}
\label{sec:experiments}

\color{black}

We begin with image classification task on the large-scale ImageNet~\cite{krizhevsky2012imagenet} dataset for studying key proprieties of the proposed method, which servers as the main benchmark dataset. Next, we use the Cityscapes~\cite{cordts2016cityscapes} dataset for image segmentation task, examining if the proposed method can also work well for dense prediction on small-scale datasets. Finally, we use the Kinetics~\cite{kay2017kinetics} dataset to demonstrate the proposed method can generalize well not only on $2$D images, but also on $3$D videos with spatial-temporal dimension for action recognition task.\footnote{Code and trained model will be released on GitHub.}

\subsection{Implementation Details}
\paragraph{Image Classification}
We first use ResNet-50~\cite{he2016identity} as a shallow CNN to conduct ablation studies and then use deeper CNNs to further exam the effectiveness of the proposed method. A variety of networks are tested as the backbone CNN, including the ResNet~\cite{he2016identity}, ResNeXt~\cite{xie2017aggregated}, Dual Path Network(DPN)~\cite{chen2017dual}, and SE-Net~\cite{hu2017}. All networks are trained with the same strategy~\cite{chen2017dual} using MXNet~\cite{chen2015mxnet} with $64$ GPUs. The learning rate is decreased by a factor of $0.1$ starting from $0.4$\footnote{For SE-Nets, we adopt $0.3$ as the initial learning rate since it diverged when using $0.4$ as the initial learning rate.}; the weight decay is set to $0.0002$; the networks are updated using SGD with a total batch size of $2,048$. We report the Top-1 
classification accuracies on the validation set with $224 \times 224$ single center crop~\cite{he2016identity,xie2017aggregated,chen2017dual}.

\paragraph{Semantic Image Segmentation}
We employ the simple yet effective Fully Convolutional Networks (FCNs)~\cite{chen2018deeplab} as the backbone. Specifically, we adopt ImageNet~\cite{krizhevsky2012imagenet} pre-trained ResNet~\cite{he2016deep}, remove the last two down-sampling operations and adopt the multi-grid~\cite{chen2017rethinking} dilated convolutions. Our proposed block(s) is randomly initialized and is appended at the end of the FCN just before the final classifier, between two adaptive convolution layers. Same with~\cite{liu2015parsenet, chen2017rethinking, chen2018deeplab}, we employ a ``poly'' learning rate policy where $power = 0.9$ and the initial learning rate is $0.006$ with batch size of $8$.

\paragraph{Video Action Recognition}
We run the baseline methods and our proposed method with the code released by \cite{chen2018multi} using PyTorch~\cite{paszke2017pytorch}. We follow \cite{wang2017non} to build the backbone $3$D ResNet-50/101 which is pre-trained on ImageNet~\cite{krizhevsky2012imagenet} classification task. However, instead of using $7\times7\times7$ convolution kernel for the first layer, we use $3\times5\times5$ convolution kernel for faster speed as suggested by \cite{chen2018a2nets}. The learning rate starts from $0.04$ and is decreased by a factor of $0.1$. Newly added blocks are randomly initialized and trained from scratch. We select the center clip with center crop for the single clip prediction, and evenly sample 10 clips per video for the video level prediction which is similar with \cite{wang2017non}.

\subsection{Results on ImageNet}
\label{sec:experiments_image}

\color{black}

We first conduct ablation studies using ResNet-50~\cite{he2016identity} as the backbone architecture and considering two scenarios: 1) when only one extra block is added; 2) when multiple extra blocks are added. We then conduct further experiments with more recent and deeper CNNs to further examine the effectiveness of the proposed unit.

\begin{figure}[t]
\centering
\resizebox{\columnwidth}{!}{
	\includegraphics[]{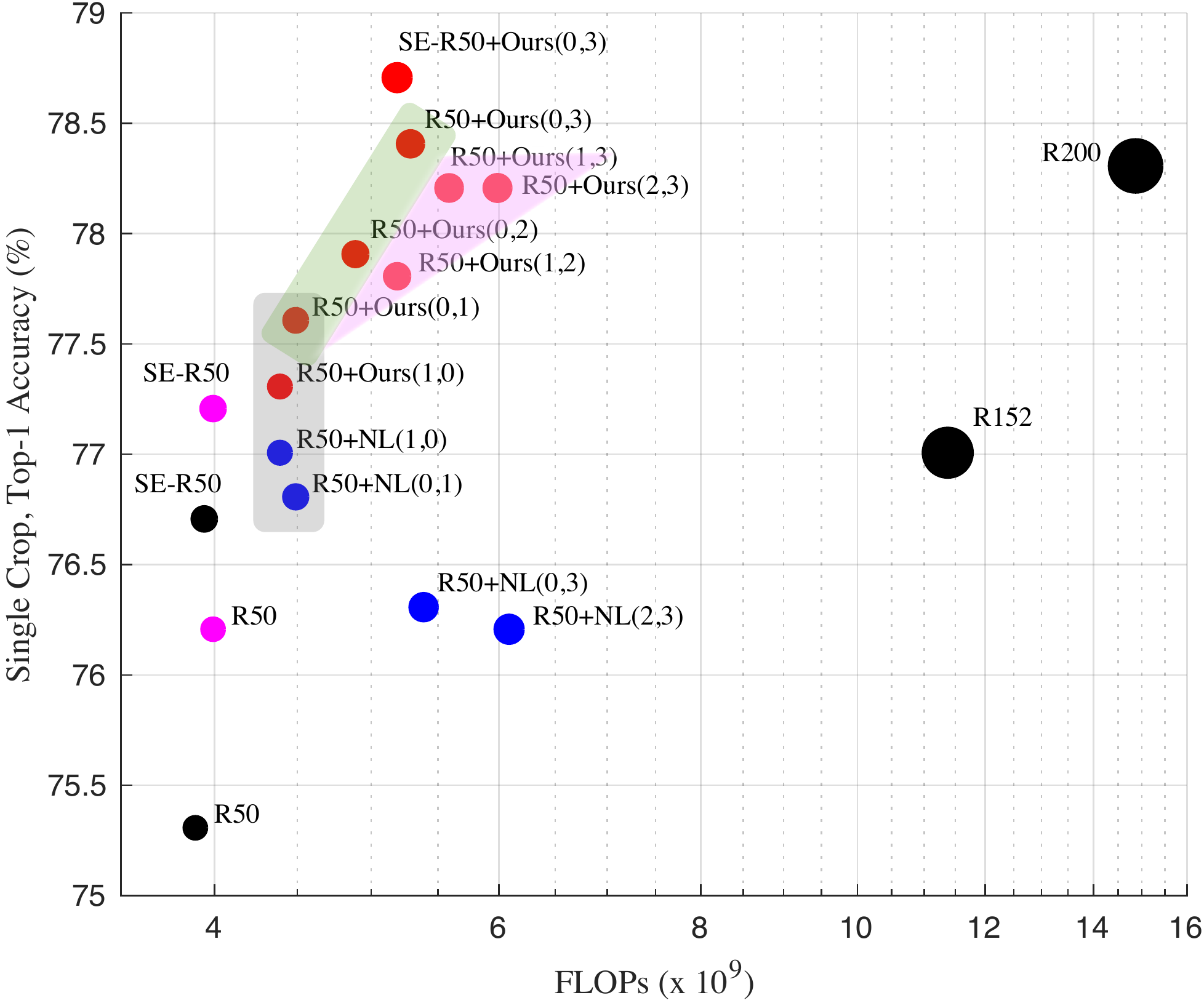}
}
\caption{
Ablation study on ImageNet validation set with ResNet-50~\cite{he2016identity} as the backbone CNN. Black circles denote results reported by authors in~\cite{he2016identity,hu2017}, while all other colors denote results reproduced by us. Specifically, red circles refer to models with at least one \oursunitshort, blue circle denote the use of the related NL unit~\cite{wang2017non}, while ``SE-'' denotes the use of SE units~\cite{hu2017}. The size of the circle reflects model size. Our reproduced ResNet-50 (R50) and SE-ResNet-50 (SE-R50) give slightly better results that reported, due to the use of strided convolution\protect\footnotemark and different training strategies.
}
\label{fig:imnet:ablation}
\end{figure}

\begin{table}[t]
\centering
\renewcommand{\arraystretch}{1.2}
\caption{Performance comparison of adding different numbers of graph convolution layers on ImageNet validation set. $g$ denotes the number of graph convolution layers inside a \oursunit. Top-1 accuracies on ImageNet validation set are reported.}
\small
\resizebox{\columnwidth}{!}{
\begin{tabular}{>{\centering}p{2cm}|>{\centering}p{1.3cm}|>{\centering}p{1.3cm}|>{\centering}p{1.3cm}|c}
  \toprule
            & \multirow{2}{*}{Plain} & \multicolumn{3}{c}{+1 \oursunitlong} \\
  \cline{3-5}
            &                        &   $g = 1$  &   $g = 2$  & ~~~$g = 3$~~~\\
  \midrule
  ResNet-50 &       76.15\%          &   77.60\%  &   77.62\%  &    77.66\%   \\
  \bottomrule
\end{tabular}
}
\label{tab:imnet:num-gcn}
\end{table}

\paragraph{Ablation Study}
Figure~\ref{fig:imnet:ablation}\footnotetext{https://github.com/facebook/fb.resnet.torch} shows the ablation study results, where the y-axis is the Top-1 accuracy and x-axis shows the computational cost measured by FLOPs, \emph{i.e.} floating-point multiplication-adds~\cite{he2016deep}. We use ``R'', ``NL'', ``Our''  to represent \emph{Residual Networks},  \emph{Nonlocal Block}~\cite{wang2017non}, our proposed method respectively, and use ``(n, m)'' to indicate insert location. For example, ``R50+Our(1,3)'' means one extra \oursunit is inserted to ResNet-50 on Res3, and three \oursunits are inserted on Res4 evenly. We first study the case when only one extra block is added as shown in gray area. Seen from the results, the proposed method improves the accuracy of ResNet-50 (pink circle) by $1.5\%$ when only one extra block is added. Compared with Nonlocal method, the proposed method shows higher accuracy under the same computation budget and model size. We also find inserting the block on Res4, \emph{i.e.} ``R50+Ours(0,1)'', gives better accuracy gain than inserting it on Res3, \emph{i.e.} ``R50+Ours(1,0)'', which is probably because Res4 contains more level features with semantics. Next, we insert more blocks on Res4 and the results are shown in the green area. We find that \oursunit can consistently lift the accuracy when more blocks are added. Surprisingly, just adding three \oursunits enhances ResNet-50 by up to $78.4\%$ in Top-1 accuracy, which is even $0.1\%$ better than the deepest ResNet-200~\cite{he2016identity}, yet with only about $30$\% GFLOPS and  $50$\% model parameters. This is very impressive, showing that our newly added block can provide some complementary features which cannot be easily captured by stacking convolution layers. Similar improvement has also been oberved on SE-ResNet-50~\cite{hu2017}. We also insert multiple blocks on different stages as shown in the purple area, and find adding all blocks at Res4 gives the best results. It is also interesting to see that the Nonlocal method starts to diverge during the optimization when more blocks are added, while we did not observe such optimization difficulties for the proposed method.\footnote{For better comparing the optimization difficulty, we do not adopt the zero initialization trick~\cite{goyal2017accurate} for both methods.} The Table~\ref{tab:imnet:num-gcn} shows the effects of using different numbers of graph convolution layers for each \oursunit. Since stacking more graph convolution layers does not give significant gain, we only use one graph convolution layer per unit unless explicitly stated.

\begin{table}[t]
\centering
\renewcommand{\arraystretch}{1.2}
\caption{Performance gain by adding our proposed \oursunit on different state-of-the-art networks on ImageNet validation set. We find the \oursunit provides consistent improvements independent of the architecture. ``+n'' means adding n extra blocks at ``Res3'' or ``Res4''.
}
\label{tab:imnet:deeper}
\resizebox{\columnwidth}{!}{
  \begin{tabular}{ll|cc|cc|c}
  \toprule
  \multicolumn{2}{c|}{ Method }                 		       & Res3  & Res4  & GFLOPs & \#Params & 		Top-1 \\
  \midrule
  \multirow{3}{*}{ResNet50~\cite{he2016identity}} 	& Baseline &       &       &   4.0  &   25.6M  &  		76.2\%  \\
    					   		            		& \oursunitshort (Ours)   &       &   +3  &   5.2  &   30.5M  &\textbf{78.4\%} \\
    					   		            		& \oursunitshort (Ours) &   +2  &   +3  &   6.0  &   31.4M  &  		78.2\%  \\
  \midrule
  \multirow{2}{*}{SE-ResNet50~\cite{hu2017}} 	    & Baseline &       &       &   4.0  &   28.1M  &  		77.2\%  \\
    					   		            		& \oursunitshort (Ours) &       &   +3  &   5.2  &   33.0M  &\textbf{78.7\%} \\
  \midrule
  \multirow{3}{*}{ResNet200~\cite{he2016identity}} 	& Baseline &       &       &  15.0  &   64.6M  &  		78.3\%  \\
    					   		            		& \oursunitshort (Ours) &       &   +3  &  16.2  &   69.7M  &  		79.4\%  \\
    					   		            		& \oursunitshort (Ours) &   +2  &   +3  &  16.9  &   70.6M  &\textbf{79.7}\% \\
  \midrule
  \multirow{2}{*}{\makecell{ResNeXt101~\cite{xie2017aggregated}\\($32\times4$)}} 
  													& Baseline &       &       &   8.0  &   44.3M  &  		78.8\%  \\
    					   		            		& \oursunitshort (Ours) &   +2  &   +3  &   9.9  &   50.3M  &\textbf{79.8}\% \\
  \midrule
  \multirow{2}{*}{DPN-98~\cite{chen2017dual}} 		& Baseline &       &       &  11.7  &   61.7M  &  		79.8\%  \\
    					   		            		& \oursunitshort (Ours)  &   +2  &   +3  &  13.6  &   67.7M  &\textbf{80.2}\% \\
  \midrule
  \multirow{2}{*}{DPN-131~\cite{chen2017dual}}    	& Baseline &       &       &  16.0  &   79.5M  &  		80.1\%  \\
    					   		            		& \oursunitshort (Ours) &   +2  &   +3  &  17.9  &   85.5M  &\textbf{80.3}\% \\
  \bottomrule
  \end{tabular}
}
\end{table}

\paragraph{Going Deeper with Our Block}
We further examine if the proposed method can improve the performance of deeper CNNs. In particular, we exam four different deep CNNs: ResNet-200~\cite{he2016identity}, ResNeXt-101~\cite{xie2017aggregated}, DPN-98~\cite{chen2017dual} and DPN-131~\cite{chen2017dual}. The results are summarized in Table~\ref{tab:imnet:deeper}, where all baseline results are reproduced by ourselves using the same training setting for fair comparison. We observe consistent performance gain by inserting \oursunit even for these very deep models where accuracies are already quite high. It is also interesting to see that adding \oursunit on both ``Res3'' and ``Res4'' can further improve the accuracy for deeper networks, which is different from the observations on ResNet-50, probably because deeper CNNs contains more informative features in ``Res3'' than the shallow ResNet-50.

\color{black}

\subsection{Results on Cityscapes}
\label{sec:experiments_segmentation}

The Cityscapes contains 5,000 images captured by the dash camera in $2048\times1024$ resolution. We use it to evaluate the dense prediction ability of the proposed method for semantic segmentation. Compared with the ImageNet, it has much fewer images with higher resolution. 
Note that we do not use the extra coarse data~\cite{cordts2016cityscapes} during training which is orthogonal to the study of our approach.

\begin{table}[t]
\centering
\renewcommand{\arraystretch}{1.2}
\caption{Semantic segmentation results on Cityscapes validation set. ImageNet pre-trained ResNet-50 is used as the backbone CNN.}
\resizebox{1.0\columnwidth}{!}{
  \begin{tabular}{cccc|cc}
  \toprule
      FCN	 &  multi-grid   & +1 \oursunit & +2 \oursunit &     mIoU    & $\Delta$ mIoU\\
  \midrule
  \checkmark &               &               &               &    75.79\%  &              \\
  \checkmark &  \checkmark   &               &               &    76.45\%  &     0.66\%   \\
  \checkmark &  \checkmark   &   \checkmark  &               &\bf{78.25\%} & \bf{2.46\%}  \\
  \checkmark &  \checkmark   &               &   \checkmark  &    77.84\%  &     2.05\%   \\
  \bottomrule
  \end{tabular}
}
\label{tab:seg:ablation}
\end{table}

The performance gain of each component is shown in Table~\ref{tab:seg:ablation}. As can be seen, adopting the multi-grid trick~\cite{chen2017rethinking} can help improve the performance, but the most significant gain comes from our proposed \oursunit. In particular, by inserting one \oursunit, the mIoU is improved by $1.8\%$ compared with the ``FCN + multi-grid'' baseline. 
Besides, we find that adding two \oursunits sequentially does not give extra gain as shown in the last row of the table.

\begin{table}[t]
\centering
\renewcommand{\arraystretch}{1.2}
\caption{Semantic segmentation results on Cityscapes test set. All networks are evaluated by the testing server. Our method is trained without using extra ``coarse'' training set.}%
\resizebox{1.0\columnwidth}{!}{
  \begin{tabular}{l|c|cccc}
  \toprule
    Method	                      		&   Backbone    &  IoU cla.  &  iIoU cla. &  IoU cat.  &  iIoU cat.  \\
  \midrule
   DeepLab-v2~\cite{chen2018deeplab}    &   ResNet101   &   70.4\%   &   42.6\%   &   86.4\%   &   67.7\%    \\
   PSPNet~\cite{zhao2017pyramid}        &   ResNet101   &   78.4\%   &   56.7\%   &   90.6\%   &   78.6\%    \\
   PSANet~\cite{jia2018psanet}          &	ResNet101	&   80.1\%   &			&			 &			     \\
   DenseASPP~\cite{yang2018denseaspp}	&	ResNet101	&   80.6\%   &			&			 &			     \\
  \midrule
   FCN + 1 \oursunit             		&   ResNet50    &   79.5\%   &   60.3\%   &   91.3\%   &   81.5\%    \\
   FCN + 1 \oursunit             		&   ResNet101   &   \textbf{80.9\% }  &   \textbf{62.2\% }  &   \textbf{91.5\%}   &   \textbf{82.1\%}    \\

  \bottomrule
  \end{tabular}
}
\label{tab:seg:testing}
\end{table}

We further run our method on the testing set and then upload its prediction to the testing server for evaluation, with results shown in Table~\ref{tab:seg:testing} along with other state-of-the-art methods. Interestingly without bells and trick (\emph{i.e.} without using extra coarse annotations, in-cooperated low-level features or ASPP~\cite{chen2017rethinking}), our proposed method that only use ResNet-50 as backbone can already achieves better accuracy than some of the popular bases, and the deep ResNet-101 based model achieves competitive performance with the state-of-the-arts.

\begin{figure*}[h]
\centering
\resizebox{1.0\textwidth}{!}{
	\includegraphics[]{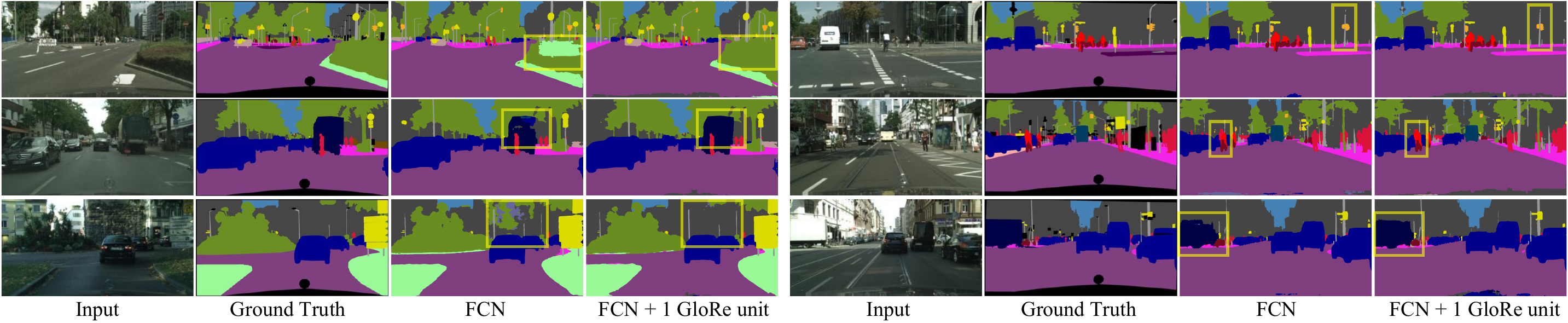}
}
\caption{
Qualitative segmentation results from the Cityscapes validation set for FCN with and without \oursunit. Differences are highlighted with yellow boxes. The figure is better viewed digitally, when zooming in.
}

\label{fig:detection:vis}
\end{figure*}

Figure~\ref{fig:detection:vis} visualizes the prediction results on the validation set. As highlighted by the yellow boxes, \oursunit enhances the generalization ability of the backbone CNN, and is able to alleviate ambiguity and capture more details.

\subsection{Results on Kinetics}
\label{sec:experiments_video}
\color{black}

The experiments presented in the previous section demonstrate the effectiveness of the propose method on 2D image related tasks. We now evaluate the performance of out \oursunit on 3D inputs and the flagship video understanding task of action recognition. We choose the large-scale Kinetics-400~\cite{kay2017kinetics} dataset fortesting that contains approximately 300k videos. We employ the ResNet-50(3D) and ResNet-101(3D) as the backbone and insert 5 extra \oursunits in total, on Res3 and Res4. The backbone networks are pre-trained on ImageNet~\cite{krizhevsky2012imagenet}, where the newly added blocks are randomly initialized and trained from scratch.

\begin{figure}[t]
\centering
\resizebox{\columnwidth}{!}{
	\includegraphics[]{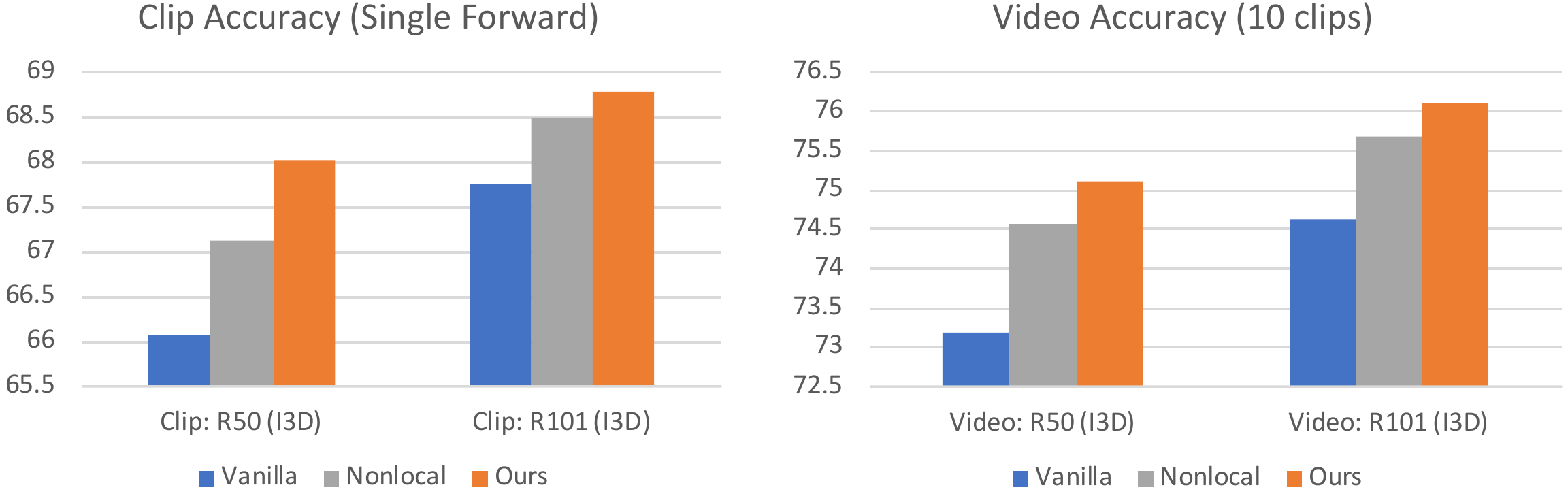}
}
\caption{Performance comparison on Kinetics-400 dataset. The clip level top-1 accuracy is shown one the left, while the video level top-1 accuracy is shown on the right.}
\label{fig:kinetics:bar}
\end{figure}

We first compare with Nonlocal Networks (NL-Net)\cite{wang2017non}, the top performing method. We reproduce the NL-Net for fair comparison since we use distributive training with much larger batch size and fewer input frames for faster speed. We note that the reproduced models
achieve performance comparable to the one
reported by authors with much lower costs. The results are shown in Figure~\ref{fig:kinetics:bar} and show that 
the proposed method consistently improves recognition accuracy over both the ResNet-50 and ResNet-101 baselines, and provides further improvement over the NL-Nets.

\begin{table}[t]
\centering
\renewcommand{\arraystretch}{1.2}
\caption{
Results on the Kinetics validation set. All methods use only RGB information (no Optical Flow).
}
\resizebox{\columnwidth}{!}{
  \begin{tabular}{l|ccc|cc}
  \toprule
  Method     							&   Backbone    &     Frames    &   FLOPs   & Clip Top-1 & Video Top-1 \\
  \midrule
  I3D-RGB~\cite{carreira2017quo}     	&  Inception-v1 &       64      & 107.9 G 	&    --      &   71.1\%   \\
  R(2+1)D-RGB~\cite{tran2017closer} 	&   ResNet-xx   &       32      & 152.4 G 	&    --      &   72.0\%   \\
  MF-Net~\cite{chen2018multi} 			&    MF-Net     &       16      &  11.1 G   &    --  	 &   72.8\%   \\
  S3D-G~\cite{xie2017rethinking}        &  Inception-v1 &       64      &  71.4 G 	&    --      &   74.7\%   \\
  \midrule
  NL-Nets~\cite{wang2017non}      	    &   ResNet-50   &        8      &  30.5 G  	&  67.12\%   &   74.57\%  \\  
  \oursunitshort (Ours)                 &   ResNet-50   &        8      &  28.9 G  	&  68.02\%   &\textbf{75.12\%}  \\
  \midrule 
  NL-Nets~\cite{wang2017non}      	    &   ResNet-101  &        8      &  56.1 G  	&  68.48\%   &   75.69 \%  \\  
  \oursunitshort (Ours)                 &   ResNet-101  &        8      &  54.5 G  	&  68.78\%   &\textbf{76.09\%} \\ 
  \bottomrule
  \end{tabular}
}
\label{tab:kinetics:all-methods}
\end{table}

All results including comparison with other prior work are shown in Table~\ref{tab:kinetics:all-methods} along with other recently proposed methods. 
Results show that by simply adding the \oursunit on basic architectures we are able to outperforms other recent state-of-the-art methods, demonstrating its effectiveness in a different, diverse task.

\section{Visualizing the GloRe Unit}
\label{sec:discussion}
\color{black}

\begin{figure}[t]
\centering
\resizebox{1.0\linewidth}{!}{
	\includegraphics[]{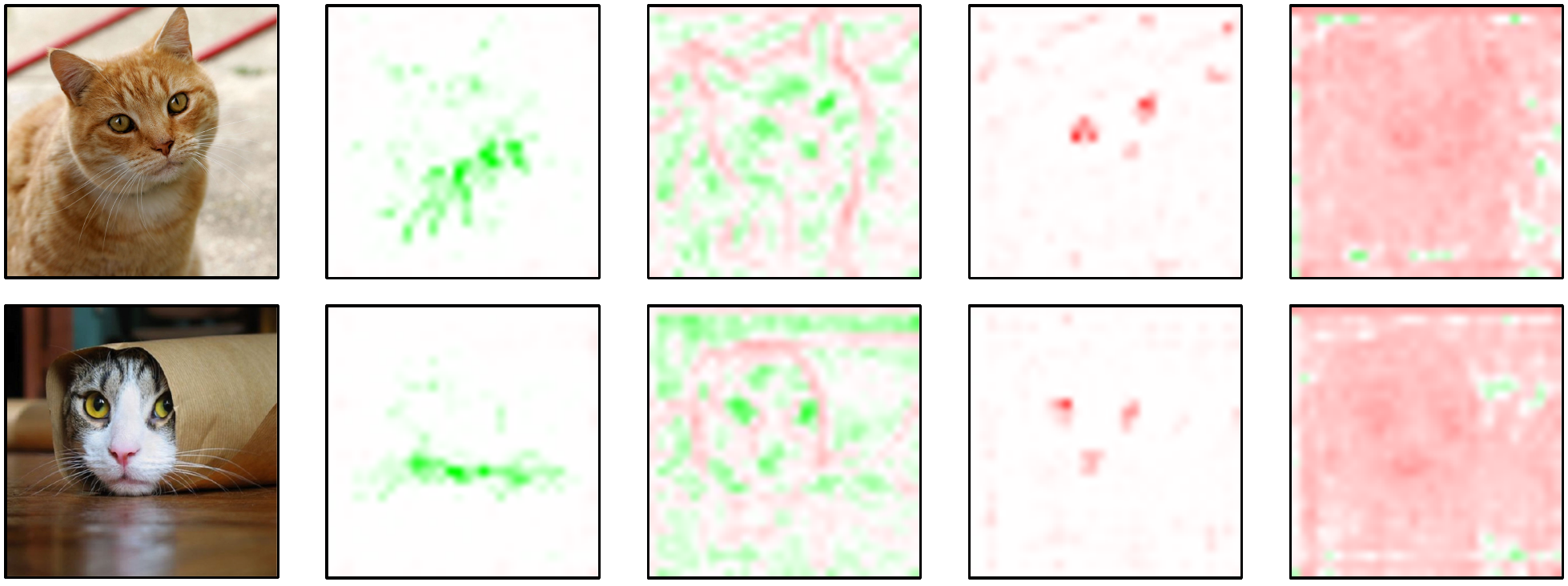}
}
\caption{Visualization of the learned projection weights (best viewed in color). Red color denotes positive and green negative values, color brightness denotes magnitude.
}
\label{fig:graph:vis}
\end{figure}

Experiments in the previous section show that the proposed method can consistently boost the accuracy of various backbone CNNs on a number of datasets for both 2D and 3D tasks. We here analyze what makes it work by visualizing the learned feature representations.

To generate higher resolution internal features for better visualization, we trained a shallower ResNet-18~\cite{he2016identity}  with one \oursunit inserted in the middle of Res4. We trained the model on ImageNet with $512 \times 512$ input crops, so that the intermediate feature maps are enlarged by $2.2\times$ containing more details. Figure~\ref{fig:graph:vis} shows the weights for four projection maps (\ie $\vb_i$ in Eqn.~\ref{eqn:global_projection}) for two images. The depicted weights would be the coefficients for the corresponding features at each location for a weighted average pooling over the whole image, giving a single feature descriptor in interaction space. For this visualization we used $N=128$ and therefore 128 such feature descriptors would be extracted for pooled regions, forming a graph with 128 nodes in interaction space. As expected, different projection weight map learn to focus on different global or local discriminative patterns. For example, the left-most weight map seems to focus on cat whiskers, the second weight maps seems to focus on edges, the third one seems to focus on eyes, and the last one focus on the entire space equally acting more like a global average pooling. As discussed in Sec~\ref{sec:introduction}, it is really hard for convolution operations to directly reason between such patterns that might be spatially distant or ill-shaped.

\section{Conclusion}

\color{black}

In this paper, we present a highly efficient approach for global reasoning that can be effectively implemented by projecting information from the coordinate space to nodes in an interaction space graph where we can directly reason over globally-aware discriminative features. The proposed \oursunit is an efficient instantiation of the proposed approach, where projection and reverse projection are implemented by weighted pooling and weighted broadcasting, respectively, and interactions over the graph are modeled via graph convolution. It is lightweight, easy to implement and optimize, while extensive experiments show that the proposed unit can effectively learn features complementary to various popular CNNs and consistently boost their performance on both 2D and 3D tasks over a number of datasets.

\clearpage
{\small
\bibliographystyle{ieee}
\bibliography{egbib}
}

\end{document}